%% file: Manuscript.tex
\documentclass[journal]{IEEEtran}
\usepackage{graphicx}
\usepackage{graphbox}

\usepackage{booktabs,graphicx}
\usepackage[table]{xcolor}
\definecolor{best}{HTML}{E0EFDF}
\definecolor{second}{HTML}{FBEFCF}

\usepackage[utf8]{inputenc}
\usepackage{url}
\usepackage{booktabs}
\usepackage{graphicx}
\usepackage{pifont}
\usepackage{booktabs}
\usepackage{pifont}
\usepackage{xcolor}
\usepackage{colortbl}

\definecolor{zjRankOneBg}{HTML}{FBE8DF}
\definecolor{zjRankOneText}{HTML}{8E3C28}
\definecolor{zjRankTwoBg}{HTML}{FFF2D7}
\definecolor{zjRankTwoText}{HTML}{76551E}
\definecolor{zjRankThreeBg}{HTML}{E8EFF8}
\definecolor{zjRankThreeText}{HTML}{3B5B83}
\newcommand{\zjRankOne}[1]{%
    \cellcolor{zjRankOneBg}{\color{zjRankOneText}\bfseries #1}}
\newcommand{\zjRankTwo}[1]{%
    \cellcolor{zjRankTwoBg}{\color{zjRankTwoText}#1}}
\newcommand{\zjRankThree}[1]{%
    \cellcolor{zjRankThreeBg}{\color{zjRankThreeText}#1}}
\newcommand{\zjRankFour}[1]{#1}
\definecolor{modelHeader}{HTML}{E8ECF0}

\definecolor{modelBaseBg}{HTML}{F6F7F8}
\definecolor{modelBaseText}{HTML}{444444}
\definecolor{modelOursBg}{HTML}{FCF0E9}
\definecolor{modelOursText}{HTML}{30343B}
\definecolor{modelGainBg}{HTML}{EDF3FA}
\definecolor{modelGainText}{HTML}{3B5B83}

\newcommand{\modelbasecell}[1]{%
    \cellcolor{modelBaseBg}{\color{modelBaseText}#1}}

\newcommand{\modelourscell}[1]{%
    \cellcolor{modelOursBg}{\color{modelOursText}#1}}

\providecommand{\gaincell}[1]{}
\renewcommand{\gaincell}[1]{%
    \cellcolor{modelGainBg}{\color{modelGainText}#1}}

\definecolor{zjHeader}{HTML}{F1F3F5}
\definecolor{zjFull}{HTML}{FCF0E9}
\definecolor{zjInk}{HTML}{30343B}
\definecolor{modelHeader}{HTML}{EEEAE6}
\definecolor{modelGainBg}{HTML}{FBF0E4}
\definecolor{modelGainText}{HTML}{99512E}
\definecolor{chrxcolor}{RGB}{1, 140, 116}

\usepackage{trimclip}
\DeclareRobustCommand{\hmark}{\rlap{\textcolor{black!25}{\ding{51}}}\rlap{\clipbox{0pt 0pt {0.5\width} 0pt}{\ding{51}}}\phantom{\ding{51}}}
\usepackage{amssymb}
\usepackage{bbding}
\usepackage{pifont}
\usepackage{algorithmicx}
\usepackage{algpseudocode}
\usepackage[nointegrals]{wasysym}
\usepackage{utfsym}
\usepackage{fontawesome}
\usepackage[dvipsnames]{xcolor}
\definecolor{c4}{HTML}{fb4c1f}
\usepackage{amsmath}

\usepackage{amssymb}

\usepackage{booktabs}
\usepackage{stfloats}
\usepackage{graphicx}
\usepackage[normalem]{ulem}
\usepackage{amsmath, amssymb, algorithm, algpseudocode}
\usepackage{amsmath}
\usepackage{bm}
\usepackage{amsfonts}
\usepackage[bookmarks=false]{hyperref}
\usepackage{colortbl}
\usepackage{booktabs}
\usepackage[per-mode=symbol,range-phrase=--,range-units=single]{siunitx}
\usepackage{cite}
\usepackage{nicefrac}
\usepackage{multirow}
\usepackage{calc}
\usepackage{bbding}
\usepackage{amssymb}
\usepackage{pifont}
\usepackage{hyperref} 
\usepackage{textcomp}
\begin{document}
\bstctlcite{IEEEexample:BSTcontrol}

\title{EarthJev: Efficient Visual Evidence Acquisition and Decision for UHR Remote Sensing Understanding}
%
%
%
\author{
    Pengyu Dai,
    Yao Zhang,~\IEEEmembership{Student Member,~IEEE,}
   Jian Song,
   Junjue Wang,
   Yafei~Ou,~\IEEEmembership{Member,~IEEE,}  
   Yuzeng Chen,
   Junshi Xia,~\IEEEmembership{Senior Member,~IEEE,} 
   Hongruixuan Chen,~\IEEEmembership{Member,~IEEE,}
   Naoto~Yokoya,~\IEEEmembership{Member,~IEEE}%
   
    \thanks{Pengyu Dai and Yao Zhang contributed equally to this work.}%
    \thanks{Pengyu Dai, Naoto Yokoya, Yafei Ou, Hongruixuan Chen ,Junshi Xia, and Jian Song are with RIKEN Center for Advanced Intelligence Project (AIP), RIKEN, Tokyo, Japan.}%
    \thanks{Yao Zhang is with the State Key Laboratory of Information Engineering in Surveying, Mapping and Remote Sensing, Wuhan University, Wuhan, China.}%
    \thanks{Junjue Wang, Pengyu Dai, Yuzeng Chen and Naoto Yokoya are with The University of Tokyo, Japan.}%
    \thanks{Corresponding author: Naoto Yokoya(yokoya@k.u-tokyo.ac.jp)}
}



\maketitle
\begin{abstract}
Ultra-high-resolution (UHR) remote sensing visual question answering (VQA) requires locating fine-grained evidence in large scenes under limited visual input budgets. Active observation addresses this challenge through selective exploration, but repeated generative judgments can make region search costly. Its decisions also concern distinct prediction targets: spatial pointer validity, evidence presence, and reasoning correctness.
We propose EarthJev, a framework that separates region scoring from answer generation and uses target-specific decision signals while keeping the answering model frozen. A spatial pointer verifier uses visual and geometric cues to guide exploration or bypass search. An evidence-aware decision mechanism scores compact candidate views in batches using a lightweight head on frozen representations, without generating textual region judgments. Format-specific confidence calibration then guides selection between candidate answers generated from memory and overview presentations of the acquired evidence.
Across five benchmarks, EarthJev achieves a mean accuracy of 53.06\% with Qwen3.5-9B, outperforming ZoomSearch with the same answering model by 2.33–14.73 percentage points per benchmark while reducing mean latency from 11.30 to 1.67 seconds per question. With four additional frozen answering models, model-specific fitting of the decision modules and calibration mappings yields mean accuracy gains of 3.41–6.53 percentage points over the corresponding single-pass baselines. 
Code can be accessed at:
\url{https://github.com/Da1daidaidai/EarthJev}
\end{abstract}

\begin{IEEEkeywords}
 Ultra-High-Resolution images understanding,  MLLMs, Decision model, Remote sensing
\end{IEEEkeywords}

%
\IEEEpeerreviewmaketitle

\section{Introduction}
%
%
%
%
\IEEEPARstart{U}{ltra}-high-resolution (UHR) remote sensing images capture both broad geographic context and fine-grained local details. requires identifying small objects, distinguishing subtle attributes, and resolving spatial relations while retaining relevant context. Multimodal large language models (MLLMs)~\cite{liu2023visual, chen2024internvl, bai2025qwen2, qwen2025qwen3vl} operate under limited visual token budgets: global downsampling obscures small targets, whereas dense tiling introduces prohibitive computational overhead and irrelevant background tokens~\cite{Luo_2025_ICCV, wang2025xlrs, wang2025geollava}. Consequently, acquiring sufficient visual evidence under a constrained budget remains a central bottleneck in UHR remote sensing visual question answering (VQA).

\begin{figure}[!t]
    \centering
    \includegraphics[width=\columnwidth]{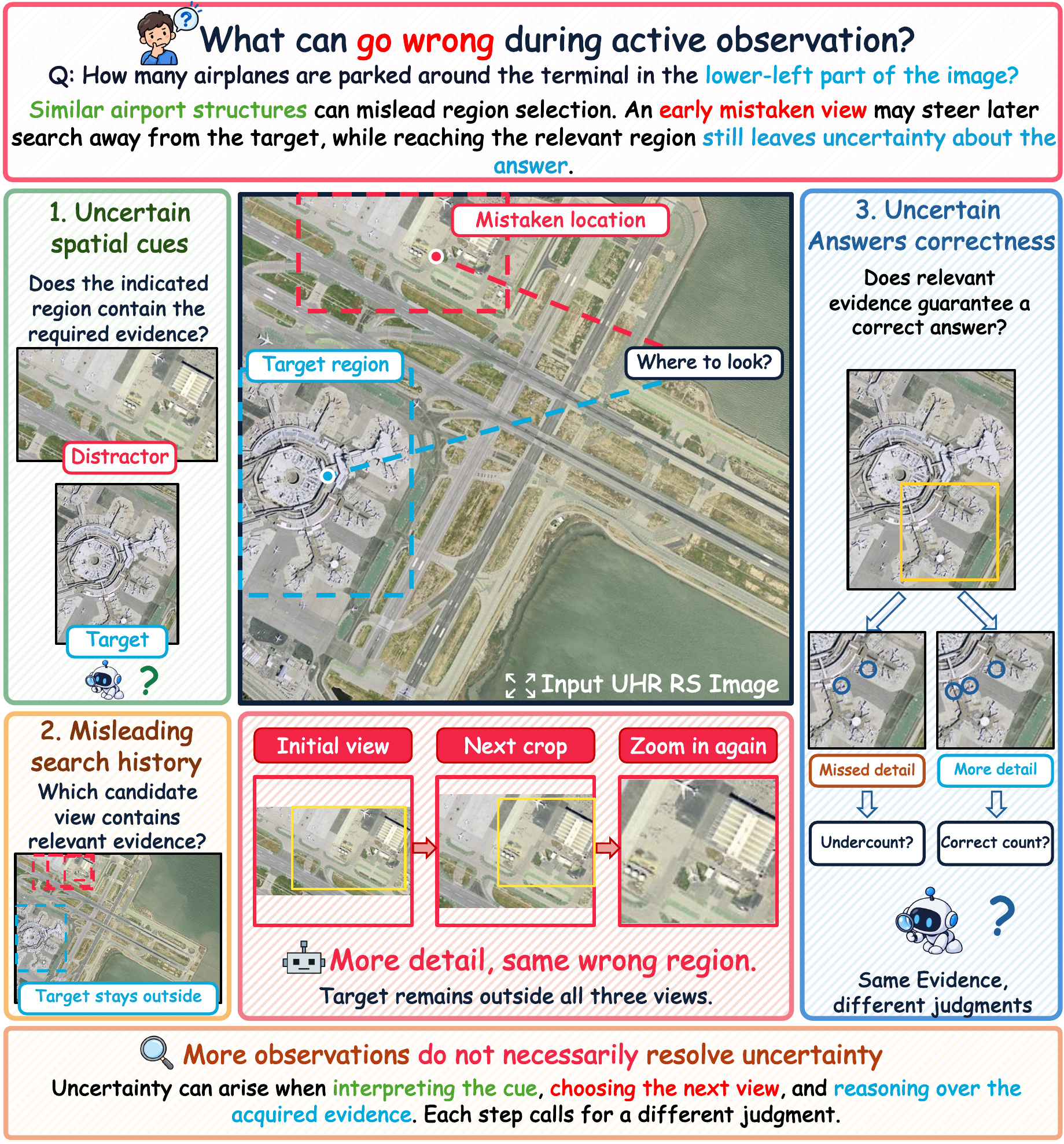}
    \caption{Challenges in active observation: ambiguous spatial cues,
    misleading search history, and uncertain answers despite relevant evidence.
    Cyan marks target-relevant regions and evidence, red indicates distractors
    and off-target search, and yellow boxes delineate crop regions.}
    \label{fig1:challenge}
\end{figure}

\begin{figure}[!t]
    \centering
    \includegraphics[width=0.99\columnwidth]{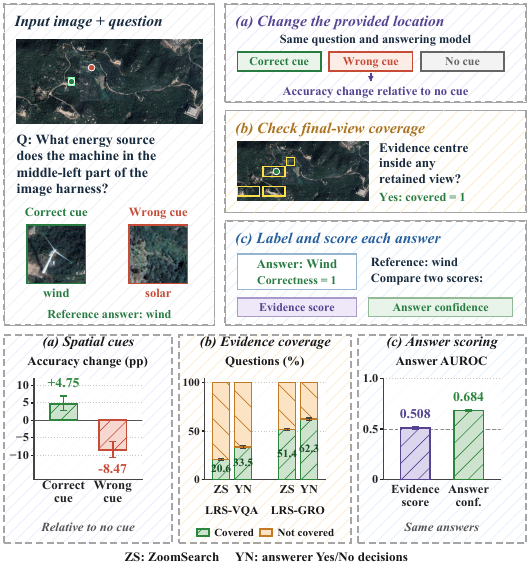}
    \caption{Preliminary diagnostics of active observation: (a) \textbf{Spatial cues:} valid cues aid perception whereas wrong cues degrade it; (b) \textbf{Evidence coverage:} heuristic search (ZS) and Yes/No probing (YN) frequently miss target evidence; and (c) \textbf{Answer scoring:} answer confidence predicts answer correctness far better than evidence scores.}

    \label{fig2:pilot}
\end{figure}

Active observation addresses this challenge through question-guided region exploration~\cite{zheng2025deepeyes, liu2026zoomearth, wang2026geoeyes, zhu2026geovista}. By selectively zooming into candidate regions guided by the question, a model can gather fine-grained details without processing the entire scene at full resolution. Yet, moving from a linguistic query to local visual evidence and finally to an answer requires three distinct judgments (Fig.~\ref{fig1:challenge}): \textit{Is the question-provided spatial pointer trustworthy? Does a candidate region contain relevant evidence? Is the answer derived from the acquired evidence reliable?} Through preliminary experiments across existing remote sensing VQA benchmarks~\cite{lobry2020rsvqa, li2024hrvqa, wang2024earthvqa, li2024vrsbench, Luo_2025_ICCV, wang2025xlrs, dang2025benchmarkultrahighresolutionremotesensing}, we find that conflating or misjudging these stages creates three critical bottlenecks:
\begin{itemize}
    \item \textbf{Spatial pointer validity:} Explicit spatial pointers, such as bounding boxes and text coordinates, provide location priors that may be inconsistent with the visual evidence. In a controlled LRS-VQA test, boxes placed on annotated evidence improve accuracy, whereas boxes placed away from it reduce accuracy below the no-box condition (Fig.~\ref{fig2:pilot}(a)).
    \item \textbf{Suboptimal region-relevance probing:} Existing search heuristics and zero-shot Yes/No token probabilities from MLLMs frequently fail to cover true evidence-bearing regions under fixed visual input (Fig.~\ref{fig2:pilot}(b)) and incur high latency when invoked repeatedly during visual region search.
    \item \textbf{Disconnect between evidence presence and answer correctness:} Locating the relevant region does not guarantee correct fine-grained perception or relational reasoning. As shown in Fig.~\ref{fig2:pilot}(c), region-level evidence-presence scores poorly separate correct from incorrect final answers, whereas calibrated answer confidence provides a much stronger decision signal.
\end{itemize}
\par These observations indicate that effective active observation cannot rely on a single undifferentiated heuristic; instead, pointer validity, evidence presence, and answer correctness should be modeled as separate, calibrated decision events.

\begin{figure*}[!t]
    \centering
    \includegraphics[width=0.99\textwidth]{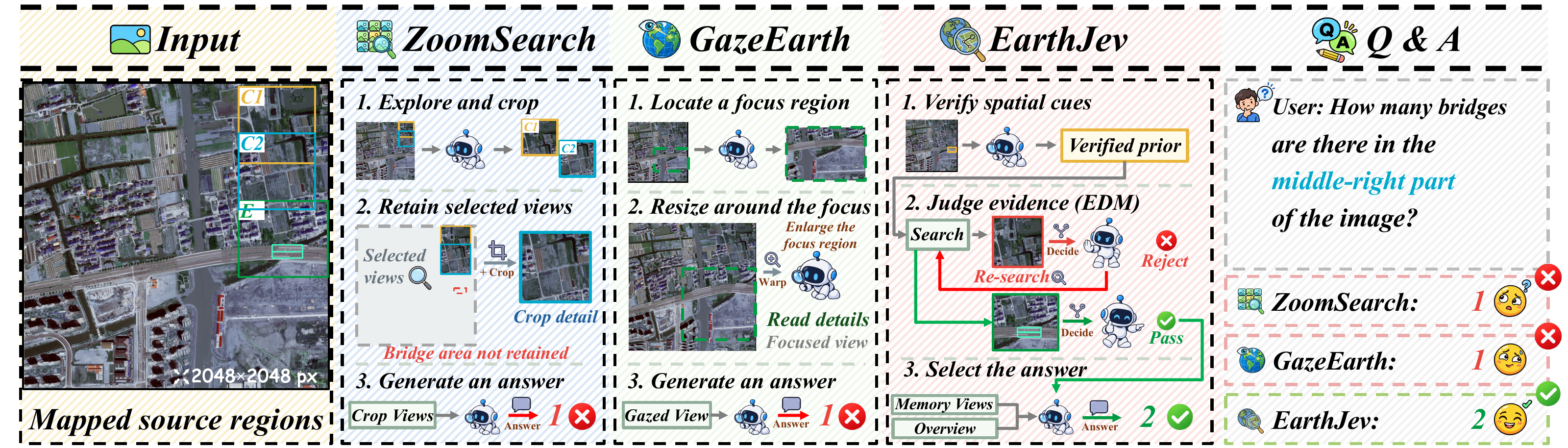}
    \caption{
Comparison of visual evidence acquisition paradigms.
ZoomSearch employs crop-based search, while GazeEarth uses non-uniform resizing. Both may focus on incomplete or misleading evidence, resulting in incorrect answers in this example.
In contrast, EarthJev verifies spatial cues, reassesses candidate views during search, and selects among candidate answers using calibrated confidence, enabling more accurate identification.
    }
    \label{fig:teaser}
\end{figure*}

Based on these insights, we propose EarthJev, a decision-enhanced framework for UHR remote sensing VQA. With the answering model kept frozen, EarthJev employs visual verification to regulate the guidance of spatial cues during the search, utilizes a lightweight region decision head to acquire relevant evidence, and selects the final answer based on independently calibrated answer confidence scores. We evaluate EarthJev on five remote sensing image benchmarks. The results demonstrate that EarthJev achieves the highest overall accuracy compared with other methods. These findings demonstrate that improving remote sensing VQA under a limited visual budget depends not only on acquiring relevant evidence but also on appropriately evaluating and utilizing that evidence, thereby translating effective observation into accurate answers(Fig.~\ref{fig:teaser}).

Our main contributions are as follows:

\begin{enumerate}
    \item \textbf{Decision targets for active observation.} We characterize the distinct roles of spatial pointer validity, evidence presence, and answer correctness through controlled pointer perturbations and diagnostic score comparisons.
    \item \textbf{Decision-Enhanced Framework:} We propose EarthJev, which integrates Spatial Pointer Verification (SPV), an Evidence-Aware Decision Mechanism (EDM), and Calibrated Answer Selection (CAS) on top of a frozen MLLM to jointly optimize where to look, when to stop, and how to select the final answer under a strict visual budget.
    \item \textbf{Superior Accuracy and Efficiency:} Extensive evaluations on five UHR remote sensing VQA benchmarks across five MLLM backbones show that EarthJev achieves state-of-the-art accuracy with 2.3–6.8$\times$ faster inference than search-based active perception baselines.

\end{enumerate}

\par The remainder of this paper is organized as follows. Section~II reviews related work on MLLMs for remote sensing VQA, active observation, and decision models. Section~III presents the proposed EarthJev framework. Section~IV describes the experimental setup and presents benchmark comparisons, ablation studies, cross-model evaluations, analyzes of the decision modules and inference efficiency, and qualitative case studies. Finally, Section~V concludes the paper.

\section{Related Works}
\subsection{MLLMs for Remote Sensing VQA}
\par Compared with natural imagery, remote sensing scenes exhibit extreme scale variations and dense distributions of small objects~\cite{ding2021object, sun2022fair1m}. Understanding such imagery requires integrating broad geographic context with local spatial relationships, placing heavy demands on both fine-grained perception and domain-specific knowledge in MLLMs~\cite{xiao2025foundation, hu2025rsgpt, luo2024skysensegpt, irvin2024teochat, soni2024earthdial, pang2024vhmversatilehonestvision, liu2024rsunivlmunifiedvisionlanguage}. To adapt MLLMs to remote sensing domains, GeoChat~\cite{kuckreja2024geochat} introduces region-level instruction tuning to ground language responses in target coordinates, while EarthGPT~\cite{zhang2024earthgpt} unifies multisensor comprehension across optical, SAR, and infrared modalities. LHRS-Bot-Nova~\cite{li2024lhrsbotnovaimprovedmultimodallarge} further enhances cross-modal alignment through multi-level visual compression and spatial instruction data, and GeoLLaVA-8K~\cite{wang2025geollava} prunes redundant background tokens while retaining anchor tokens to accommodate higher-resolution inputs in a single pass. While these methods strengthen domain representations and token encoding efficiency, they rely on costly end-to-end backbone fine-tuning and static single-pass compression. 

\par In contrast, EarthJev keeps the answering MLLM completely frozen and reformulates UHR understanding as an active, multi-stage decision process, allowing lightweight decision modules to dynamically allocate a compact visual budget and generalize across diverse MLLM backbones.

\subsection{Active Observation for UHR Remote Sensing VQA}
\par Since question-relevant targets in UHR remote sensing imagery often occupy only a tiny fraction of the scene, active perception~\cite{jiang2026vlm, gemini-twi} recasts VQA as a sequential observation process that progressively reaches into informative local regions through region selection, zooming, and other visual operations. One line of work trains MLLMs via supervised fine-tuning (SFT) or reinforcement learning (RL) to invoke cropping and visual tools~\cite{zheng2025deepeyes, lai2025minio3scalingreasoningpatterns, wang2025pixelreasonerincentivizingpixelspace, wang2026text, zhu2026geovista, jiang2026rsopsdreliableprivilegedonpolicyselfdistillation}. For instance, ZoomEarth~\cite{liu2026zoomearth} optimizes region-guided zooming trajectories via RL, GeoEyes~\cite{wang2026geoeyes} incorporates evidence-gain rewards to regulate on-demand zooming and stopping, and GeoLens~\cite{wang2026zoominglearningmultitoolvisual} integrates multi-tool execution into policy training.

\par A complementary line of work explores training-free active observation over frozen MLLMs~\cite{zhang2025mllmsknowlooktrainingfree, yu2025zoomrefineboostinghighresolutionmultimodal, dalal2026constructivedistortionimprovingmllms, ma2026weaveearthstructuredevidenceconstruction}. Specifically, ZoomSearch~\cite{zhou2025lookmatterstrainingfreeultrahr} combines patch–text relevance and model-evidence scores to guide adaptive multi-branch search, then reassembles selected patches into a layout-preserving canvas, whereas GazeEarth~\cite{zhang2026earthgazetrainingfreeactive} applies non-uniform gaze-based warping to emphasize salient regions while preserving peripheral context. 
WeaveEarth~\cite{ma2026weaveearthstructuredevidenceconstruction} constructs spatially complementary evidence sets and organizes local views with spatial metadata for reasoning.
Despite their effectiveness, existing paradigms remain susceptible to misleading spatial cues and uncalibrated exploration (as illustrated in Fig.~\ref{fig:teaser}): blindly following imprecise spatial pointers in the query or relying on zero-shot heuristic scores can steer crop-based search (e.g., ZoomSearch) or non-uniform warping (e.g., GazeEarth) toward incomplete or irrelevant regions.

\par Unlike these approaches, EarthJev explicitly verifies spatial pointers against visual content before searching, i.e., routing reliable cues to a fast path while down-weighting erroneous ones, and replaces expensive autoregressive region probing with batched hidden-state scoring to achieve both higher localization accuracy and lower latency.

\subsection{Decision Model}

\par Decision-oriented models restrict predictions to task-defined output spaces so that their scores can directly support selection, routing, and verification. Jev~\cite{almeida2026jev} provides a typed decision interface that returns probability estimates in parallel without free-form text generation. NanoJev~\cite{chen2026nanojev} implements candidate scoring on a shared language-model backbone and supports choice, Boolean, and ordered-score outputs. Visual Jev~\cite{yu2026visualjevaccurateefficient} extends this interface to multiple questions sharing an image, combining answer-supervised adaptation with candidate-token readout from the existing language-model head and batched execution over a shared visual prefix. These approaches motivate constrained probability readout and efficient execution for repeated decisions.

Efficient score extraction and probability calibration address different aspects of decision making. Guo et al.~\cite{guo2017calibrationmodernneuralnetworks} show that accurate neural classifiers can still produce poorly calibrated confidence estimates. Post-hoc calibration fits mappings from model scores to probability estimates using labeled calibration data while keeping the underlying predictor fixed. Platt scaling fits a sigmoid mapping to decision scores~\cite{platt1999probabilistic}, whereas temperature scaling rescales logits using a fitted scalar~\cite{guo2017calibrationmodernneuralnetworks}. Calibration concerns agreement between predicted probabilities and empirical outcome frequencies and must be interpreted with respect to the prediction target and the distribution represented by the calibration data.

For answer selection, the probability assigned to a generated answer is not automatically an estimate of its correctness. Kadavath et al.~\cite{kadavath2022language} study candidate-level self-evaluation by eliciting the probability that a proposed answer is correct. Tian et al.~\cite{tian2023justask} further show that confidence-elicitation strategies affect calibration in language models fine-tuned with human feedback. In the visual setting, Visual Jev~\cite{yu2026visualjevaccurateefficient} reports that an auxiliary evidence-sufficiency head detects missing evidence but does not improve answer accuracy or selective prediction in its evaluated setting. These findings motivate distinguishing evidence-related predictions from answer-correctness estimates when choosing a signal for a downstream decision. Recently, JEV-style decision-making methods have also emerged in the RS domain~\cite{rsjev,eojev}, but most overlook the computationally expensive UHR tasks.

\par Moving beyond single-step Jev formulations, EarthJev decouples and calibrates three stage-specific decision signals across the entire perception-to-reasoning loop, culminating in a Calibrated Answer Selection (CAS) module that reconciles complementary evidence presentations (\textit{memory} vs. \textit{canvas}) to ensure that acquired visual evidence reliably translates into correct answers.

\begin{figure*}[!t]
    \centering
    \includegraphics[width=0.99\textwidth]{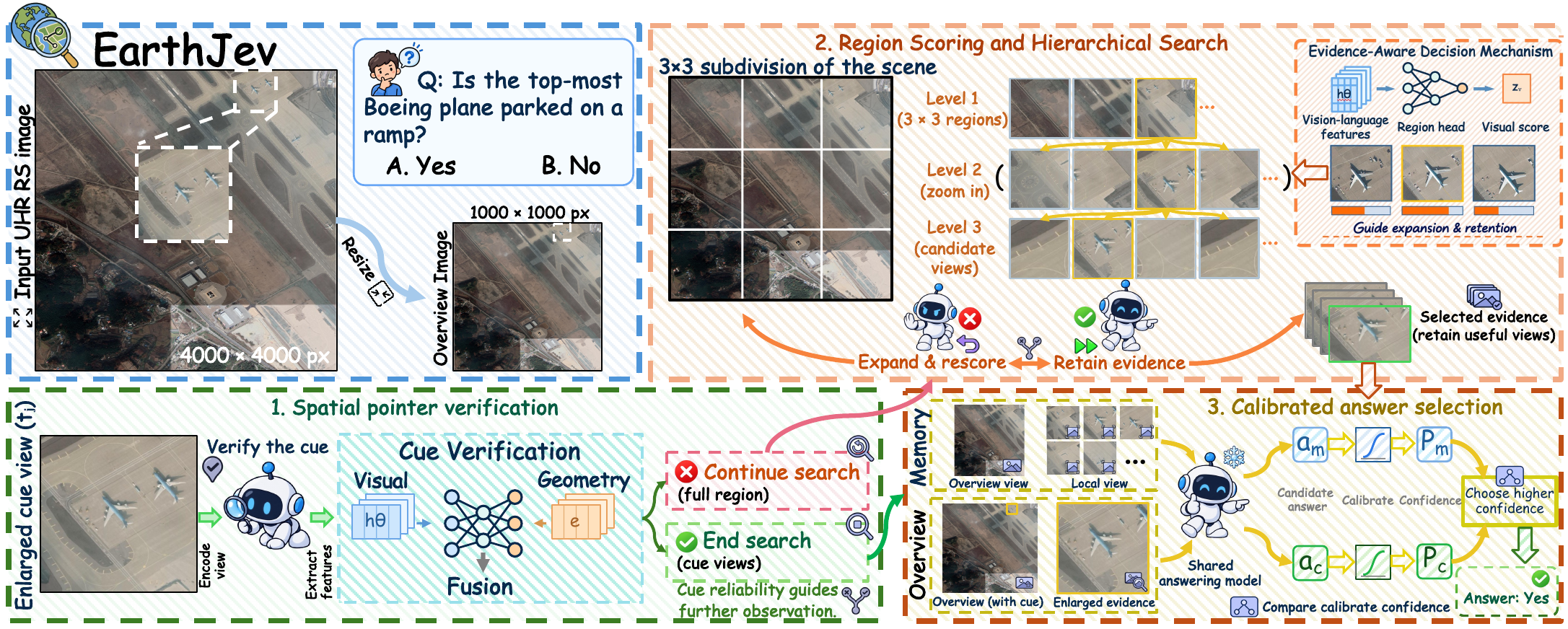}
    \caption{
The EarthJev workflow consists of three stages: (1) Spatial pointer verification, which integrates visual and geometric information to evaluate the localization cues provided in a question, determining whether the perspective presented by these cues is sufficient to answer the question directly or if further searching is required. (2) An EDM, which utilizes vision-language representations to score candidate regions, thereby guiding hierarchical expansion and the retention of evidence. (3) The selected evidence is organized into a ``memory'' and a ``canvas'', which are then accessed by a shared, frozen answering model. Finally, the system evaluates candidate answers based on calibrated confidence scores and selects the most reliable one as the definitive answer.
    }
    \label{fig:framework}
\end{figure*}

\section{Methods}
\subsection{Problem Formulation and Framework Construction}
Given an UHR remote sensing image \(I\) and a natural language question \(q\), our objective is to generate an answer \(\hat{a}\) using a frozen multimodal answering model \(\mathcal{A}_{\theta}\) under a limited visual budget. Here, \(q\) consists of the question text and optional candidate answers. Rather than directly feeding the entire image into the model, EarthJev progressively selects local views based on the question, acquiring the details necessary for answering while preserving global scene information. As shown in Fig.~\ref{fig:teaser}, let \(\mathcal{V}=\{c_i\}_{i=1}^{m}\) denote the set of selected views. This process can be formulated as
\begin{equation}
\begin{aligned}
\mathcal{V} &= \mathcal{S}(I,q;\mathcal{B}),\\
\hat{a} &= \mathcal{A}_{\theta}\left(q,\mathcal{T}(I,\mathcal{V})\right),
\end{aligned} 
\label{eq1}
\end{equation}
where \(\mathcal{S}\) denotes the evidence selection process, \(\mathcal{B}\) represents the constraints on search and visual input budgets, and \(\mathcal{T}\) organizes the global scene and local views into the input of the answering model. Accordingly, solving the question involves two interconnected stages: locating relevant evidence within large-scale imagery and utilizing the acquired evidence to generate reliable answers.
\begin{equation}
\label{eq2}
\mathcal{V}=\mathcal{S}(I,q,\mathcal{R}(q);g_\psi,f_\phi,\mathcal{B}),
\end{equation}
Here, \(K\) denotes the maximum number of retained views. When the question does not contain an explicit spatial cue, \(\mathcal{R}(q)=\varnothing\). Through this process, the spatial cues provided in the question and the visual feedback obtained during search jointly determine subsequent observations, allowing evidence acquisition to progress incrementally based on the currently available information. Even after relevant regions have been identified, the model may still generate incorrect answers due to errors in fine-grained perception or the understanding of spatial relationships. Therefore, EarthJev further evaluates candidate answers after evidence selection, rather than directly reusing the region scores from the search stage. Specifically, EarthJev organizes the same set of evidence into two presentation formats, memory and canvas, to generate candidate answers \(a_r\), and calibrates their respective raw confidence features \(z_r\):
\begin{equation}
\begin{aligned}
a_r &= \mathcal{A}_{\theta}
\left(q,\mathcal{T}_{r}(I,\mathcal{V})\right),\\
P_r &= \mathcal{C}_{\eta_r}(z_r),\\
r^{*} &= \underset{r\in\{\mathrm{memory},\mathrm{canvas}\}}{\arg\max}\;P_r,
\qquad \hat{a}=a_{r^{*}},
\end{aligned} 
\end{equation}
where \(\mathcal{C}_{\eta_r}\) denotes the calibration function corresponding to each evidence presentation format, and \(P_r\) estimates the probability that the candidate answer is correct. In this way, spatial cue verification establishes the starting point for observation, region-level decisions guide evidence search, and answer confidence determines the final output, collectively forming a complete process from questions to evidence and ultimately to answers. Throughout the framework, the parameters \(\theta\) of the answering model remain frozen, with only lightweight decision modules trained and the corresponding calibration parameters fitted.

\subsection{Spatial Pointer Verification}
\label{sec:spv}
In the evidence selection process of Eq.~\ref{eq2}, the spatial
cue set \(\mathcal{R}(q)\) provides initial locations for
region exploration. Whether these locations contain the
required evidence, however, depends on their visual content.
EarthJev therefore uses a verifier \(g_{\psi}\) to assess
the indicated regions and determine how strongly the cues
should guide subsequent observations. For each
\(r_j\in\mathcal{R}(q)\), we map its coordinates to the
original image and extract a local view \(t_j\) that
preserves the surrounding context. The verifier combines
its visual score with geometric and option-structure
features to estimate pointer reliability:
\begin{equation}
\label{eq:spv}
\begin{aligned}
p_j
&= g_{\psi}(t_j,q,\mathcal{R}(q))\\
&= \sigma\!\left(
\frac{z_v(t_j)+\gamma_j\mathbf{w}^{\top}\mathbf{e}_j}{T_g}
+b_{\mathrm{pos}}(t_j)
\right).
\end{aligned}
\end{equation}
Here, \(z_v(t_j)\) is computed by the region decision head
\(f_{\phi}\) from frozen answering-model representations,
and \(\mathbf{e}_j\) encodes 22-dimensional geometric and
option-structure features through an MLP.
The gate \(\gamma_j=\sigma(\mathbf{w}_g^{\top}[\mathbf{e}_j;z_v(t_j)]+b_g)\) modulates their contribution, \(T_g\) is the temperature of the verifier, and \(b_{\mathrm{pos}}\) encodes directional priors. We train the verifier with cross-entropy on
annotation-derived pointers and corrupted counterparts
generated by shifting, rescaling, or relocating them.

The reliability assessed at the pointer views \(t_j\)
then determines how much the search relies on these
locations. For each candidate search view \(c\), we use
\(\pi=\max_j p_j\) to interpolate between scores with
and without explicit coordinate information:
\begin{equation}
    p_{\mathrm{mix}}(c)
=\pi\,p(c\mid\mathcal{R}(q))
+(1-\pi)\,p(c\mid\mathcal{R}_{\mathrm{pos}}(q)).
\end{equation}
The first term applies the same verifier, enabling
geometric fusion for views associated with explicit
pointers. The second disables this fusion while retaining
visual evidence and directional priors
\(\mathcal{R}_{\mathrm{pos}}(q)\); it is used alone when
no explicit pointer is available. Thus, stronger visual
support at the indicated locations gives pointer geometry
greater influence over subsequent exploration.
The resulting \(p_{\mathrm{mix}}(c)\) guides node expansion
and evidence retention in the following section. Beyond guiding search, the estimated pointer reliability routes the question. For this routing decision,
we combine \(p_j\) with the frozen answering model's
affirmative Yes/No probability \(y_j\) for evidence
relevance in the same view \(t_j\).
Their mean logit,
\begin{equation}
P_{\mathrm{ptr}}
=\sigma\!\left(
a_{\mathrm{ptr}}\max_j\ell_j+b_{\mathrm{ptr}}
\right).   
\end{equation}
The parameters \(a_{\mathrm{ptr}}\) and \(b_{\mathrm{ptr}}\) are fitted using pointer-validity labels.
When \(P_{\mathrm{ptr}}\geq\tau_{\mathrm{fast}}\), EarthJev bypasses search and answers from the pointer view with the highest \(p_j\). Otherwise, it performs hierarchical search and merges the pointer views with the views it retains.
For questions with bounding boxes only in the answer
options, the fast path instead uses
\(\max_j p_j\geq\tau_{\mathrm{opt}}\).

\subsection{Region Scoring and Hierarchical Search}
When spatial cue verification does not trigger the fast path, EarthJev employs hierarchical search to acquire the local evidence required for answering. The key challenge in this process is determining which views warrant further exploration. To this end, we develop EDM that predicts the presence of relevant evidence within a region using the model's vision-language representations and incorporates this prediction into region expansion and evidence retention. 
For each candidate view \(c\), we resize it to at most
256 visual tokens and pair it with an evidence-querying
prompt derived from \(q\). The resulting hidden
representation \(\mathbf{h}_{\theta}(c,q)\) is passed
to a two-layer decision head \(f_{\phi}\) to estimate
whether the view contains question-relevant evidence.
Explicit coordinates are excluded from the prompt and
handled by the verification mechanism in Section~\ref{sec:spv},
while directional descriptions provide a soft positional
prior. The resulting region score is as follows:
\begin{equation}
p(c\mid q)
=
\sigma\!\left[
\frac{f_{\phi}(\mathbf{h}_{\theta}(c,q))}{T}
-\sum_{a\in\{x,y\}}
s_a\left|u_a(c)-u_a^{\star}\right|
\right].
\end{equation}
Here, \(u_a(c)\) denotes the position of the view center along axis \(a\), \(u_a^{\star}\) represents the target position corresponding to the directional description, and \(s_a\) controls the strength of the prior. The region decision head is trained using binary cross-entropy loss on cropped samples with annotated evidence regions, while \(\theta\) remains frozen throughout training. The temperature \(T_f\) is fitted on held-out samples. Consequently, each view can be assigned a region score through a single representation extraction, without requiring the generation of textual responses for region-level judgments.
Specifically, starting from the full image, each expansion divides the
selected view into nine subregions, whose scores are computed
in one batch. For questions with explicit spatial pointers,
we use the fused score \(p_{\mathrm{mix}}(c)\) from Section~\ref{sec:spv};
otherwise, we use \(p(c\mid q)\).
Denoting the applicable score by \(p_{\mathrm{search}}(c)\),
we select the next view from the set of expandable
candidate nodes \(\mathcal{F}\):
\begin{equation}
c^{*}
=
\underset{c\in\mathcal{F}}{\arg\max}\;
p_{\mathrm{search}}(c).
\end{equation}
After each expansion, the newly generated subregions are added to the candidate set, and the next node is selected based on the region scores until the search expansion budget is exhausted. Region-level judgments thus directly determine the observation order, guiding the search progressively toward locations more likely to contain question-relevant evidence. After the search is completed, evidence is selected relative to the average score of the candidate views. Let \(\bar p_{\mathrm{search}}\) denote the mean score of the candidate views. We retain up to six views satisfying \(\sigma(2[p_{\mathrm{search}}(c)-\bar p_{\mathrm{search}}])>0.6\); if no view satisfies this condition, the highest-scoring view is retained. These views are merged with the pointer views. Each view \(c\) is assigned an evidence presence probability
\begin{equation}
\label{eq:pres}
P_{\mathrm{pres}}(c)=\sigma\!\left(a_{\mathrm{pres}}\operatorname{logit}p(c)+b_{\mathrm{pres}}\right),
\end{equation}
where \(p(c)\) is the search score \(p_{\mathrm{search}}(c)\) or, for a pointer view, its verifier score \(p_j\), and \((a_{\mathrm{pres}},b_{\mathrm{pres}})\) is fitted by Platt scaling. Each view is enlarged to include context (1.5 times its size and at least 512 pixels per side), the views are ranked by \(P_{\mathrm{pres}}(c)\), a view whose IoU with a higher-ranked view exceeds 0.5 is merged into it, and the top \(K\) views form \(\mathcal{V}\).

\subsection{Calibrated Answer Selection}
After obtaining the evidence set \(\mathcal{V}\), EarthJev organizes the visual input to the answering model in two formats: memory sequentially presents a global thumbnail and local views with positional information, while canvas assembles the same set of local views into a unified canvas according to their spatial layout~\cite{zhou2025lookmatterstrainingfreeultrahr}, as shown in Fig.~\ref{fig:visual_memory}. The answering model processes the two input formats separately to generate the candidate answers \(a_r\) in Eq. (3). Given the same evidence, the framework selects the final answer based on the confidence scores of the candidate answers. Specifically, we extract the confidence feature \(z_r\) from the answer generation probabilities. For multiple-choice questions, we use the predicted probability of the selected option letter and apply a logit transformation. For open-ended questions, we use the average log-probability of the answer tokens:
\begin{equation}
z_r=
\begin{cases}
\operatorname{logit}(u_r),
& \text{MCQ},\\[4pt]
\displaystyle
\frac{1}{L_r}\sum_{t=1}^{L_r}
\log p_{\theta}\!\left(
a_{r,t}\mid a_{r,<t},q,\mathcal{T}_r(I,\mathcal{V})
\right),
& \text{open-ended}.
\end{cases} 
\end{equation}
Here, \(u_r\) denotes the predicted probability of the selected option letter, and \(L_r\) represents the number of tokens in the candidate answer. Since the relationship between raw confidence scores and answer correctness may differ across evidence presentation formats, we fit separate Platt scaling mappings for each benchmark (RSHR-Bench, which provides no calibration questions, reuses the mappings fitted on MME-RS and XLRS-Bench-lite) to convert
:
\begin{equation}
P_r
=\mathcal{C}_{\eta_r}(z_r)
=\sigma(\alpha_r z_r+\beta_r).
\end{equation}
Here, \(\eta_r=(\alpha_r,\beta_r)\) is fitted separately for answer correctness under each evidence presentation format. EarthJev then compares the calibrated confidence scores \(P_r\) according to Eq. (3) and selects the candidate answer with the highest confidence as the final answer \(\hat{a}\). Finally, evidence presence vetoing corrects answers that contradict the retrieved evidence. If the selected answer is an option stating that the queried object does not appear in the image, while \(\max_{c\in\mathcal{V}}P_{\mathrm{pres}}(c)\geq 0.5\), EarthJev replaces it with the remaining option of the highest predicted probability. In this way, EarthJev connects question-guided active observation with evidence-based answering, enabling more accurate and efficient remote sensing VQA under a limited visual budget.

\input{table/comparison_sota}

\section{Experiments}

\subsection{Experimental Setup}
We evaluate EarthJev on the complete question sets of the five benchmarks described below (25,039 questions in total). Because the relation between answer confidence and correctness may differ with the answer format and scoring rule of each benchmark, the calibration mappings and decision thresholds of EarthJev are fitted on a small labeled calibration set, which consists of 460 questions sampled from MME-RS, LRS-VQA, and XLRS-Bench-lite and of the official training split of LRS-GRO (Section~\ref{sec:impl}).
\subsubsection{Benchmarks}
\par We evaluate EarthJev on the complete evaluation splits of five UHR remote sensing VQA benchmarks spanning diverse spatial scales and reasoning skills:
\begin{itemize}
    \item \textbf{LRS-VQA}~\cite{Luo_2025_ICCV} comprises 7,333 open-ended questions across three subsets, i.e. LRS-FAIR~\cite{sun2022fair1m}, LRS-Bridge~\cite{glhbridge}, and LRS-STAR~\cite{star}, evaluating object attributes, counting, and scene comprehension.
    \item \textbf{MME-RealWorld-RS (MME-RS)}~\cite{zhang2025mmerealworldmultimodalllmchallenge} contains 3,738 single-choice questions targeting fine-grained color recognition, object counting, and position identification.
    \item \textbf{LRS-GRO}~\cite{liu2026zoomearth} organizes 17 question categories across Global, Region, and Object levels with region-of-interest annotations. In this study, we evaluate on its test set and report the results separately for the Global, Region, and Object groups; its official training split is used only for calibration.
    \item \textbf{RSHR-Bench}~\cite{dang2025benchmarkultrahighresolutionremotesensing} assesses UHR perception and reasoning; we evaluate on its 1,154-question single-turn benchmark subset.
    \item \textbf{XLRS-Bench-lite (XLRS-lite)}~\cite{wang2025xlrs} includes 3,080 multiple-choice questions across 13 fine-grained Level-3 (L3) task categories.
\end{itemize}

\subsubsection{Compared Baselines}
\par Corresponding to the four horizontal groups in Table~\ref{tab:main1} and Table~\ref{tab:main2}, we compare EarthJev against 21 representative methods:
\begin{itemize}
    \item \textbf{Training-Free Active Observation:} Region cropping and multi-branch search methods including ViCrop~\cite{zhang2025mllmsknowlooktrainingfree}, ZoomSearch~\cite{zhou2025lookmatterstrainingfreeultrahr}, Zoom-Refine~\cite{yu2025zoomrefineboostinghighresolutionmultimodal}, and WeaveEarth~\cite{ma2026weaveearthstructuredevidenceconstruction}, as well as non-uniform spatial warping methods including AttWarp~\cite{dalal2026constructivedistortionimprovingmllms} and GazeEarth~\cite{zhang2026earthgazetrainingfreeactive}.
    \item \textbf{Trained Remote Sensing Active Perception:} Domain-adapted zooming and tool-use agents trained via supervised fine-tuning, self-distillation, or reinforcement learning, including ZoomEarth~\cite{liu2026zoomearth}, GeoEyes~\cite{wang2026geoeyes}, GeoLens~\cite{wang2026zoominglearningmultitoolvisual}, and RS-OPSD~\cite{jiang2026rsopsdreliableprivilegedonpolicyselfdistillation}.
    \item \textbf{General-Domain Visual Search Agents:} Reinforcement-learned thinking-with-images models designed for general high-resolution reasoning, including DeepEyes~\cite{zheng2025deepeyes}, Mini-o3~\cite{lai2025minio3scalingreasoningpatterns}, and Pixel-Reasoner~\cite{wang2025pixelreasonerincentivizingpixelspace}.
    \item \textbf{Single-Pass Remote Sensing MLLMs:} End-to-end instruction-tuned remote sensing models that process the whole image in a single forward pass, including GeoChat~\cite{kuckreja2024geochat}, VHM~\cite{pang2024vhmversatilehonestvision}, RSUniVLM~\cite{liu2024rsunivlmunifiedvisionlanguage}, SkySenseGPT~\cite{luo2024skysensegpt}, TEOChat~\cite{irvin2024teochat}, EarthDial~\cite{soni2024earthdial}, GeoLLaVA-8K~\cite{wang2025geollava}, and Coarse-to-Fine~\cite{Luo_2025_ICCV}.
\end{itemize}
\par All baselines are evaluated using their official codebases and default prompting strategies. ZoomSearch and Zoom-Refine share the same frozen Qwen3.5-9B backbone as EarthJev, other training-free methods use Qwen3-VL-8B, and trained methods retain the backbones of their released checkpoints.

\subsubsection{Evaluation Metrics}
\par For a question set $\mathcal{Q}$, we define
\begin{equation}
\mathrm{Acc}(\mathcal{Q})
=\frac{100}{|\mathcal{Q}|}\sum_{i\in\mathcal{Q}}\delta_i,
\qquad \delta_i\in\{0,1\},
\label{eq:exp_accuracy}
\end{equation}
where $\delta_i$ is determined by the benchmark-specific answer matcher.
We use the official MME-RealWorld and XLRS-Bench parsers for multiple-choice
answers. XLRS-Bench-lite includes multi-answer land-use questions, for
which the predicted option set must match the reference set. For
RSHR-Bench, which supplies no scoring script in the evaluation setup, we
extract the leading option letter or identify the option whose full text
is reproduced. Open-ended matching follows the LRS-VQA and ZoomEarth
evaluation scripts, including lowercasing and WordNet~\cite{miller-1994-wordnet} synonym matching.
Failed or empty responses receive $\delta_i=0$.

\par Each benchmark's Avg. is question-level accuracy. Subset and task scores use
the corresponding question sets. For RSHR-Bench, Perception and Reasoning
are unweighted means over their respective task categories, while its Avg.
is accuracy over all evaluated questions; these quantities use different
weighting rules. Overall Avg. gives equal weight to the five benchmarks:
\begin{equation}
\mathrm{Acc}_{\mathrm{overall}}
=\frac{1}{|\mathcal{D}|}\sum_{b\in\mathcal{D}}
\mathrm{Acc}(\mathcal{Q}_b),\qquad |\mathcal{D}|=5.
\label{eq:exp_overall_accuracy}
\end{equation}
Accuracy differences are reported in percentage points (pp).

For the binary decision diagnostics, we report AUROC for discrimination and binary top-label expected calibration error (ECE)~\cite{2888116.2888120} for calibration. For an event probability $p_i$ and label $o_i$, let
$\kappa_i=\max(p_i,1-p_i)$ and
$\omega_i=\mathbb{I}[\mathbb{I}[p_i\geq0.5]=o_i]$.
Using $M=15$ equal-width confidence bins,
\begin{equation}
\mathrm{ECE}
=\sum_{m:\,|B_m|>0}\frac{|B_m|}{N}
\left|\operatorname{acc}(B_m)-\operatorname{conf}(B_m)\right|,
\label{eq:exp_ece}
\end{equation}
where $\operatorname{acc}(B_m)$ and $\operatorname{conf}(B_m)$ are the bin means of $\omega_i$ and $\kappa_i$, respectively. The diagnostic outcomes are evidence presence, pointer validity, and answer correctness. ECE is evaluated on held-out questions that are not used to fit the corresponding mapping. The 460 calibration questions of MME-RS, LRS-VQA and XLRS-Bench-lite are part of the evaluation sets; excluding them changes the accuracy of EarthJev by at most 0.09 percentage points (63.16\%, 34.81\% and 55.90\% on the remaining questions).

\input{table/performance_XLRS-Bench-lite}

\subsubsection{Implementation Details}
\label{sec:impl}
\par EarthJev keeps the answering model frozen throughout, training only the region decision head and the spatial cue verification module. The lightweight decision modules are trained using the ZwZ-RL-VQA dataset~\cite{wei2026zoomingzoomingregiontoimagedistillation}, from which 12,000 questions and their corresponding region annotations are selected to construct training samples, with the proportion of incorrect spatial cues set to 25\%. During inference, a hierarchical region partitioning strategy of \(3 \times 3\) is adopted, with a maximum of 10 search expansions. The input for each candidate view is limited to 256 visual tokens, and the subregions of the same node are evaluated in batches. The memory-based evidence presentation retains at most four local views. The fast-path threshold for spatial cue verification is set to 0.60, and answer confidence scores under the two evidence presentation strategies are calibrated separately using Platt scaling. In addition, we select 90, 240, and 130 annotated questions from MME-RS, LRS-VQA, and XLRS-Bench-lite, respectively, along with 3,511 questions from the official training split of LRS-GRO, to fit the calibration parameters. This calibration process does not update the answering model. The compared methods retain their respective prompting strategies and visual processing pipelines, and evaluation on each benchmark follows its official scoring protocol.

\begin{figure}[!t]
    \centering
    \includegraphics[width=0.99\columnwidth]{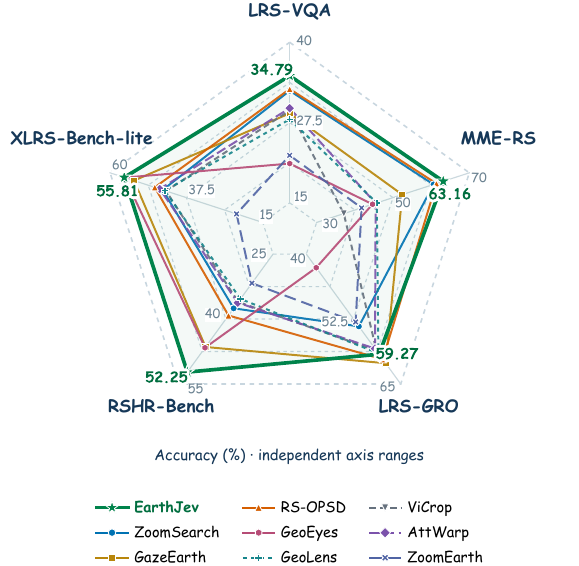}
    \caption{Comparison of overall accuracy across five remote sensing benchmarks.}
    \label{fig: radar}
\end{figure}

\subsection{Comparison with State-of-the-Art Methods}
Table~\ref{tab:main1} and Fig.~\ref{fig: radar} summarizes the comparative results of EarthJev and existing methods on five remote sensing VQA benchmarks. Using Qwen3.5-9B~\cite{qwen2026qwen35} as the answering model, EarthJev achieves an average accuracy of 53.06\% across the five benchmarks, outperforming the strongest competing method, GazeEarth, by 4.70 percentage points. It also achieves the highest overall accuracy on LRS-VQA, MME-RS, RSHR-Bench, and XLRS-Bench-lite. Specifically, EarthJev achieves accuracies of 34.79\% and 63.16\% on LRS-VQA and MME-RS, respectively, surpassing the best competing results on both benchmarks. On the evaluation subset of RSHR-Bench, EarthJev achieves an accuracy of 52.25\%, exceeding GeoEyes by 5.63 percentage points. These results demonstrate that integrating spatial cue verification, region-level evidence assessment, and answer selection throughout the active observation process can effectively improve overall performance across diverse remote sensing VQA scenarios.

\input{table/ablation_study}

A closer examination of the task-level results shows that EarthJev achieves the best performance on all three subsets of LRS-VQA, namely FAIR, Bridge, and STAR, as well as on the three task categories of MME-RS, including color recognition, object counting, and position identification. This demonstrates its effectiveness across different image contents and fine-grained perception requirements. Table~\ref{tab:main2} presents the results for the 13 L3 categories of XLRS-Bench-lite. EarthJev achieves an overall accuracy of 55.81\% and attains the best performance in the OCC, OMS, and RLUC categories, with accuracies of 61.88\%, 80.00\%, and 88.00\%, respectively. In particular, its accuracy in the OMS category exceeds the strongest competing result by 8.33 percentage points. These findings indicate that the benefits of EarthJev extend across various local visual understanding tasks, supporting the effectiveness of question-guided evidence acquisition and answer evaluation in improving high-resolution image understanding.

Table~\ref{tab:answering_models} further investigates the applicability of EarthJev across different answering models. Compared with the baselines that directly process the entire image using the same answering models, incorporating EarthJev improves the average accuracy across the five benchmarks by 6.53 and 4.99 percentage points for Qwen3.5-4B and Qwen3-VL-8B~\cite{qwen2025qwen3vl}, respectively. For Intern-S1-mini~\cite{bai2025interns1scientificmultimodalfoundation} and LLaVA-OneVision-2-8B~\cite{an2026llavaonevision2nextgenerationperceptualintelligence}, the corresponding average improvements are 3.41 and 6.52 percentage points, respectively. Moreover, all four models achieve positive improvements on each of the five benchmarks. Specifically, Qwen3.5-4B achieves improvements of 12.01 and 13.34 percentage points on XLRS-Bench-lite and RSHR-Bench, respectively, while LLaVA-OneVision-2-8B achieves an improvement of 18.59 percentage points on LRS-GRO. With the answering model parameters kept frozen, these results demonstrate the applicability of EarthJev across different model families and parameter scales. They also indicate that improving the decision-making process between observation and answering can further leverage the remote sensing understanding capabilities of existing multimodal models.

\subsection{Ablation Study and Cross-Model Generalization}
\subsubsection{Component-Wise Ablation}
\par To investigate the effects of progressively incorporating different modules, we use Qwen3.5-9B as the answering model and compare four incremental configurations presented in the Table~\ref{tab:component_ablation}. Group 1 employs the baseline search and answering pipeline, achieving an average accuracy of 45.80\% across the five benchmarks. After introducing EDM, Group 2 improves the average accuracy to 47.24\%, with improvements observed on four of the five benchmarks, indicating that region scoring based on evidence presence can improve search results. With the further incorporation of SPV, Group 3 achieves an average accuracy of 52.64\%. In particular, the accuracies on XLRS-Bench-lite and RSHR-Bench increase by 10.71 and 15.16 percentage points, respectively, demonstrating the effectiveness of spatial cue verification and its guidance of subsequent observations.

\par The complete configuration, Group 4, further incorporates calibrated answer selection while retaining evidence presence vetoing, achieving the highest average accuracy of 53.06\%, an improvement of 7.26 percentage points over Group 1. Compared with Group 3, the complete configuration yields further improvements on MME-RS, XLRS-Bench-lite, and LRS-GRO, with accuracy on LRS-GRO increasing from 56.04\% to 59.27\%. This result indicates that answer evaluation after evidence acquisition can provide additional benefits. The progressively increasing average performance across the four configurations supports the design of integrating region-level assessment, spatial cue verification, and answer selection into a unified observation pipeline.

\subsubsection{Generalization across Answering Models}
\par As shown in Tab.~\ref{tab:answering_models}, to assess generality across answering models, we also evaluate EarthJev with Qwen3.5-4B and Qwen3-VL-8B~\cite{qwen2025qwen3vl} from the Qwen family, as well as Intern-S1-mini~\cite{bai2025interns1scientificmultimodalfoundation} and LLaVA-OneVision-2-8B~\cite{an2026llavaonevision2nextgenerationperceptualintelligence} from other families. The results demonstrate that EarthJev can serve as a plug-and-play module, easily generalizing across different types of MLLM backbones and delivering consistent performance improvements.

\input{table/cross_model_generalization}

\subsection{In-Depth Analysis of Decision Modules and Efficiency}
\begin{table}[!t]
\centering
\caption{Answer selection over the same retrieved evidence with Qwen3.5-9B (accuracy, \%). Raw confidence and CAS
choose between the memory and canvas answers by raw or calibrated confidence. }
\label{tab:cas_analysis}
\begingroup
\fontsize{7.5}{9}\selectfont
\setlength{\tabcolsep}{2pt}
\renewcommand{\arraystretch}{1.15}
\resizebox{\columnwidth}{!}{%
\begin{tabular}{lcccccc}
\toprule
\textbf{Selection} & \textbf{MME-RS} & \textbf{LRS-VQA} & \textbf{XLRS-lite} & \textbf{LRS-GRO} & \textbf{RSHR-Bench} & \textbf{Avg.} \\
\midrule
Memory only & 62.92 & 33.15 & 55.00 & 58.53 & \textbf{54.16} & 52.75 \\
Canvas only & 56.18 & \textbf{35.20} & 53.64 & 56.04 & 45.84 & 49.38 \\
Raw confidence & 62.17 & 34.53 & \textbf{56.01} & 58.88 & 51.82 & 52.68 \\
\rowcolor{zjFull}
CAS (EarthJev) & \textbf{63.16} & 34.79 & 55.81 & \textbf{59.27} & 52.25 & \textbf{53.06} \\
\bottomrule
\end{tabular}
}
\endgroup
\end{table}

\begin{figure*}[!t]
\centering
\includegraphics[width=0.9\textwidth]{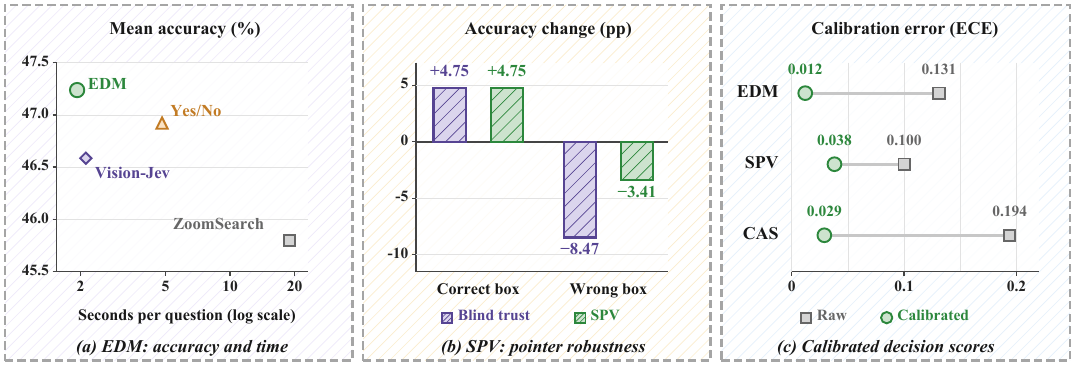}
\caption{Design evidence for the three modules (Qwen3.5-9B). (a)~EDM vs.\ other decision sources: mean accuracy over
the five benchmarks against seconds per question. (b)~SPV vs.\ blind trust: accuracy change on the LRS-VQA pointer test
with a correct or a wrong stated box, relative to no box. (c)~Held-out calibration error of each module's decision
score (evidence presence, pointer trust, answer confidence) before and after Platt scaling.}
\label{fig:memory}
\end{figure*}

\input{table/efficiency}

Table~\ref{tab:cas_analysis} compares answer-selection rules over the same retrieved evidence. CAS achieves a mean accuracy of 53.06\%, compared with 52.75\% for memory only, 49.38\% for canvas only, and 52.68\% for raw-confidence selection. Thus, its average improvement is approximately 0.3 percentage points over memory only and 0.4 percentage points over raw confidence. The gain is not uniform: CAS trails memory only on RSHR-Bench and raw-confidence selection on XLRS-Bench-lite. These benchmark-dependent results support CAS as an additional refinement for choosing between evidence presentations.

\begin{figure*}[!h]
\centering
\includegraphics[width=0.99\textwidth]{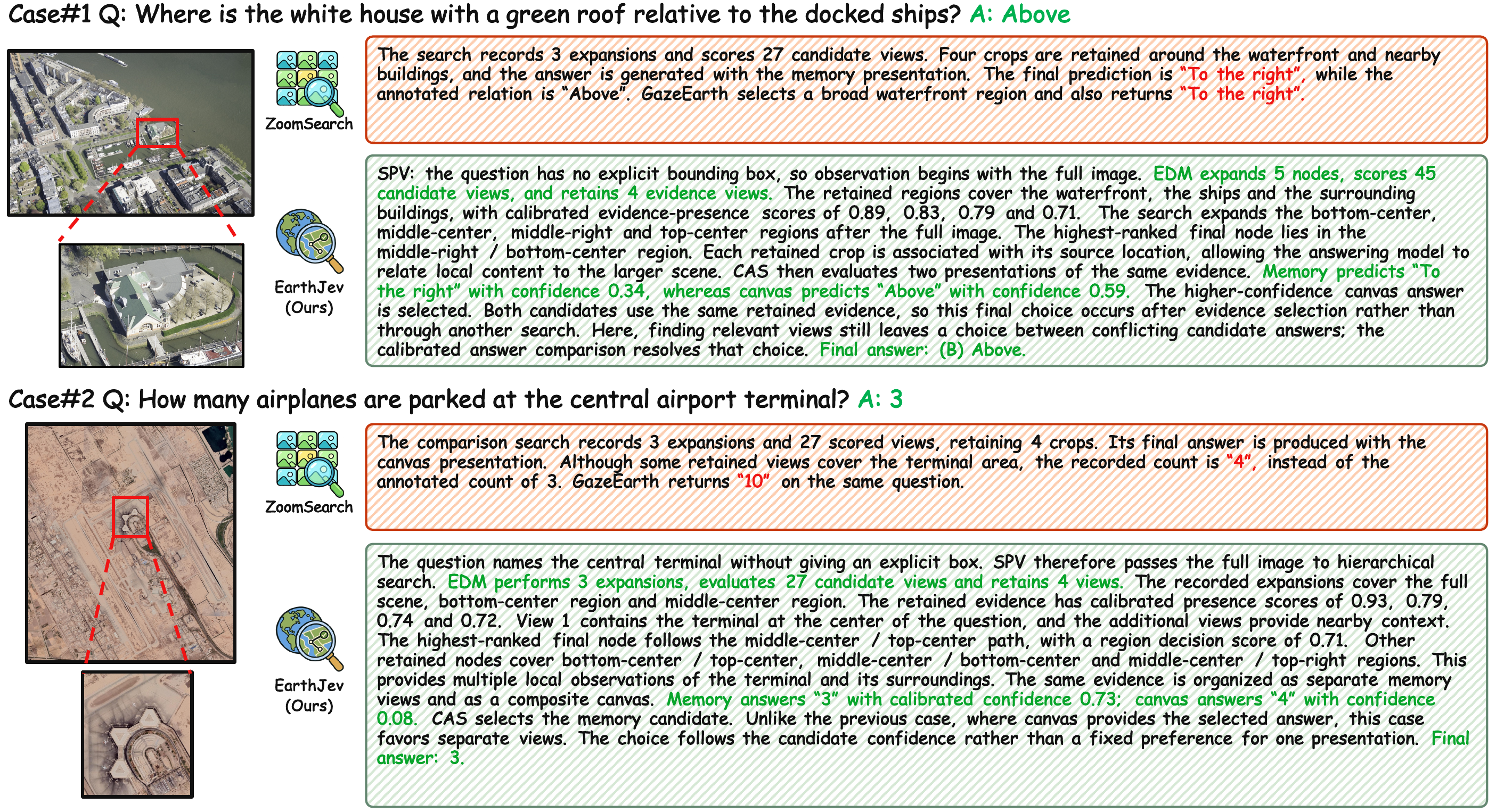}
\caption{Qualitative comparison on spatial reasoning and object counting. EarthJev selects the correct answer by evaluating calibrated confidence across memory and canvas presentations of the retrieved evidence.}
\label{fig:case}
\end{figure*}

\subsubsection{Region Probing Discrimination}
\par Fig. \ref{fig:memory}(a) compares the accuracy and inference time of different region decision methods. In this search configuration comparison, EDM achieves an average accuracy of 47.24\% across the five benchmarks, with an average inference time of 1.93 seconds per question. In contrast, using Yes/No judgments from the answering model yields an average accuracy of 46.92\%, with an inference time of 4.81 seconds per question. Combined with the region probing results in Fig. \ref{fig2:pilot}(b), the evidence recognition AUROC of the region decision head increases from 0.752 to 0.787 under the same compact-view and batch-processing conditions, while the inference time per decision remains comparable. These two sets of results demonstrate that the dedicated region decision head provides more accurate evidence scoring, while the search pipeline combining compact views and batched evaluation achieves a better accuracy–latency trade-off.

\subsubsection{Robustness to Inaccurate Spatial Pointers}
Fig. \ref{fig:memory}(b) further investigates observation behavior when spatial cues are inaccurate. In the controlled experiments on LRS-VQA, directly using correct spatial cues improves accuracy by 4.75 percentage points, whereas using incorrect cues leads to a decrease of 8.47 percentage points. After incorporating SPV, the improvement from correct spatial cues remains at 4.75 percentage points, while the accuracy degradation caused by incorrect cues is reduced to 3.41 percentage points. These results indicate that visual verification preserves the benefits of valid spatial cues while mitigating the interference of incorrect locations with subsequent searches, allowing spatial information in questions to serve as a verifiable and adjustable basis for search.

\subsubsection{Probability Calibration across Decision Stages}
\par Fig. \ref{fig:memory}(c) presents the calibration performance of different decision scores. On their respective held-out evaluation samples, the expected calibration errors (ECEs)~\cite{2888116.2888120} of evidence presence scores and spatial cue reliability scores decrease from 0.131 and 0.100 to 0.012 and 0.038, respectively. Meanwhile, the average ECE of answer confidence scores across the five benchmarks decreases from 0.194 to 0.029. Calibration improves the agreement between predicted probabilities and the empirical frequencies of their corresponding events, providing a basis for threshold-based decisions and candidate answer comparisons. Together with the different discriminative capabilities of evidence presence and answer confidence for answer correctness shown in Fig. \ref{fig2:pilot}(c), these results support constructing and calibrating separate decision signals for determining whether a region contains relevant evidence, whether a spatial cue is reliable, and whether a candidate answer is correct.

\subsubsection{Efficiency Analysis}
\par Table~\ref{tab:efficiency} compares inference latency within each answering model. With Qwen3.5-9B, EarthJev reduces mean latency from 11.30 s/q for ZoomSearch and 11.34 s/q for Zoom-Refine to 1.67 s/q, corresponding to speedups of approximately 6.8×. With Qwen3-VL-8B, it achieves 50.92\% mean accuracy at 1.68 s/q, compared with 48.36\% at 3.88 s/q for GazeEarth. These comparisons support a favorable accuracy–latency trade-off relative to the evaluated search methods. The single-pass baselines remain faster, at 3.30 and 3.41 s/q, respectively.

\subsection{Qualitative Case Study}
\par Fig.~\ref{fig:case} presents two examples illustrating how EarthJev connects evidence retrieval with answer selection. In the spatial reasoning example, the question asks for the position of a white house with a green roof relative to the docked ships. ZoomSearch and GazeEarth both answer ``To the right.'' EarthJev retrieves local views covering the waterfront, buildings, and ships, while preserving their source locations. With this evidence, the memory presentation also produces ``To the right,'' whereas the canvas presentation produces the correct answer, ``Above.'' CAS assigns higher calibrated confidence to the canvas candidate and selects it as the final response.

\par The aircraft-counting example exhibits the opposite selection. ZoomSearch predicts four airplanes, and GazeEarth predicts ten, while the annotated count is three. After EDM retrieves views of the central terminal and its surroundings, EarthJev produces a count of three from memory and four from canvas. CAS selects the correct memory candidate based on its higher calibrated confidence. Together, these examples show that the preferred evidence presentation can vary across questions. EDM supplies relevant local observations, while CAS distinguishes between the answers derived from them, helping translate evidence acquisition into accurate spatial reasoning and object counting. Fig.~\ref{fig:visual_memory} further illustrates the evidence provided to the answering model. The global overview preserves scene context, while the indexed local views expose details
of the selected regions and retain their correspondence to the original image. The examples cover object identification, energy-source recognition, color recognition, and counting. Memory and over produce the same correct answer in the first three cases, but differ in the aircraft-counting case, where CAS selects the correct memory response.

\section{Conclusion}
We proposed EarthJev, a framework that reformulates active observation in UHR remote sensing VQA as a sequence of calibrated, target-specific decisions over a frozen MLLM. Our study reveals that effective visual evidence acquisition under a limited token budget requires strictly distinguishing among three prediction targets: whether a linguistic spatial pointer is trustworthy, whether a local region contains question-relevant evidence, and whether a candidate answer derived from that evidence is reliable. By addressing these targets through SPV, a batched EDM, and CAS, EarthJev eliminates the need for expensive autoregressive region judgments and end-to-end backbone fine-tuning. Comprehensive experiments on five UHR remote sensing benchmarks verify that EarthJev substantially improves both VQA accuracy and inference efficiency, generalizes seamlessly across diverse MLLM architectures, and remains robust against perturbed spatial cues. More broadly, our results demonstrate that the primary bottleneck in applying MLLMs to gigapixel-scale Earth observation lies not only in visual encoding capacity but also in the calibration of intermediate and final decision signals. Decoupling lightweight, efficient decision modules from frozen foundation models offers a practical paradigm for resource-efficient remote sensing intelligence.

\bibliographystyle{IEEEtran}
\bibliography{IEEEabrv,cas-refs}

\end{document}

%% file: table/comparison_sota.tex
\begin{table*}[!t]
\centering
\caption{Comparison of EarthJev with existing methods across five remote sensing VQA benchmarks. Results are reported as accuracy (\%). The top three distinct scores per column are highlighted in coral, amber, and blue; best results are boldfaced, and ties share the same color. Benchmark Avg. denotes question-level accuracy, while Overall Avg. is the mean across five benchmarks. RSHR-Bench Perception/Reasoning scores are averaged over task categories.}

\label{tab:main1}

\fontsize{7.5}{9}\selectfont
\setlength{\tabcolsep}{2pt}
\renewcommand{\arraystretch}{1.15}

\resizebox{\textwidth}{!}{%
\begin{tabular}{llccccccccccccccccc}
\toprule
Method & Backbone
& \multicolumn{4}{c}{LRS-VQA}
& \multicolumn{4}{c}{MME-RS}
& \multicolumn{4}{c}{LRS-GRO}
& \multicolumn{3}{c}{RSHR-Bench}
& XLRS-lite & Overall \\

\cmidrule(lr){3-6}
\cmidrule(lr){7-10}
\cmidrule(lr){11-14}
\cmidrule(lr){15-17}
\cmidrule(lr){18-18}
\cmidrule(lr){19-19}

& & FAIR & Bridge & STAR & Avg.
& Color & Count & Position & Avg.
& Global & Region & Object & Avg.
& Perception & Reasoning & Avg.
& Avg. & Avg. \\

\midrule

ZoomSearch~\cite{zhou2025lookmatterstrainingfreeultrahr}
& Qwen3.5-9B
& \zjRankTwo{29.75} & 31.26 & \zjRankFour{34.31} & \zjRankThree{32.46}
& \zjRankThree{70.20} & \zjRankTwo{39.31} & \zjRankFour{71.84} & \zjRankThree{60.62}
& 67.14 & 48.52 & 50.31 & 53.91
& 35.74 & 55.97 & 37.52
& 44.48 & 45.80 \\

GazeEarth~\cite{zhang2026earthgazetrainingfreeactive}
& Qwen3-VL-8B
& \zjRankThree{29.18} & 26.74 & 29.38 & 28.94
& 63.67 & 21.78 & 71.12 & 52.43
& \zjRankThree{77.96} & 50.39 & \zjRankTwo{58.04} & \zjRankTwo{61.00}
& \zjRankThree{44.49} & 59.27 & \zjRankThree{46.45}
& \zjRankThree{52.95} & \zjRankTwo{48.36} \\

ViCrop~\cite{zhang2025mllmsknowlooktrainingfree}
& Qwen3-VL-8B
& 27.33 & 29.57 & 30.51 & 29.39
& 46.85 & 16.97 & 47.02 & 37.11
& 75.56 & \zjRankFour{51.48} & 54.03 & 58.58
& 34.58 & 61.67 & 36.05
& 43.51 & 40.93 \\

AttWarp~\cite{dalal2026constructivedistortionimprovingmllms}
& Qwen3-VL-8B
& 26.76 & 31.54 & 30.96 & 29.74
& 54.58 & 19.00 & 62.13 & 45.45
& 77.75 & 49.00 & 53.39 & 58.18
& 33.80 & 60.34 & 36.22
& 45.36 & 42.99 \\

WeaveEarth~\cite{ma2026weaveearthstructuredevidenceconstruction}
& Qwen3-VL-8B
& 27.64 & 24.95 & \zjRankTwo{34.78} & 31.15
& 54.02 & 23.49 & 65.71 & 47.94
& 75.34 & 47.74 & \zjRankThree{56.50} & 58.94
& 32.74 & \zjRankFour{63.76} & 36.14
& 41.36 & 43.10 \\

Zoom-Refine~\cite{yu2025zoomrefineboostinghighresolutionmultimodal}
& Qwen3.5-9B
& 28.61 & \zjRankFour{31.73} & 34.11 & \zjRankFour{32.06}
& \zjRankFour{69.96} & \zjRankFour{34.99} & \zjRankThree{75.10} & \zjRankFour{60.22}
& \zjRankOne{79.38} & \zjRankThree{51.79} & \zjRankOne{58.79} & \zjRankOne{62.06}
& 32.74 & \zjRankOne{69.52} & 35.18
& 45.19 & \zjRankFour{46.94} \\

\midrule

RS-OPSD~\cite{jiang2026rsopsdreliableprivilegedonpolicyselfdistillation}
& Qwen3-VL-8B
& \zjRankFour{28.96} & \zjRankThree{34.18} & \zjRankThree{34.48} & \zjRankTwo{32.73}
& \zjRankTwo{70.28} & \zjRankThree{36.95} & \zjRankTwo{76.29} & \zjRankTwo{61.37}
& \zjRankTwo{78.26} & \zjRankOne{53.18} & 55.21 & \zjRankThree{60.24}
& \zjRankFour{37.51} & \zjRankTwo{66.01} & \zjRankFour{39.17}
& \zjRankFour{46.82} & \zjRankThree{48.07} \\

GeoLens~\cite{wang2026zoominglearningmultitoolvisual}
& Qwen2.5-VL-7B
& 24.21 & 30.32 & 29.56 & 28.01
& 53.23 & 19.74 & 63.64 & 45.75
& \zjRankFour{77.92} & 47.47 & \zjRankFour{55.73} & 59.09
& 33.08 & 54.29 & 35.36
& 43.86 & 42.41 \\

GeoEyes~\cite{wang2026geoeyes}
& Qwen2.5-VL-7B
& 20.64 & 20.81 & 21.48 & 21.12
& 51.55 & 27.57 & 54.42 & 44.65
& 67.91 & 30.84 & 36.38 & 42.61
& \zjRankTwo{46.89} & 63.73 & \zjRankTwo{46.62}
& \zjRankTwo{55.68} & 42.14 \\

ZoomEarth~\cite{liu2026zoomearth}
& Qwen2.5-VL-3B
& 19.50 & 27.02 & 22.81 & 22.39
& 48.45 & 26.10 & 50.36 & 41.76
& 66.37 & 43.60 & 51.25 & 53.06
& 30.26 & 52.76 & 31.72
& 22.73 & 34.33 \\

\midrule

DeepEyes~\cite{zheng2025deepeyes}
& Qwen2.5-VL-7B
& 6.12 & 8.00 & 9.70 & 8.35
& 56.57 & 27.57 & 64.92 & 49.87
& 32.69 & 6.14 & 8.41 & 13.68
& 32.21 & 56.92 & 35.79
& 42.69 & 30.08 \\

Mini-o3~\cite{lai2025minio3scalingreasoningpatterns}
& Qwen2.5-VL-7B
& 12.85 & 12.24 & 13.53 & 13.13
& 52.83 & 23.16 & 60.54 & 45.69
& 57.09 & 25.52 & 28.08 & 34.42
& 18.91 & 27.28 & 20.36
& 31.53 & 29.03 \\

Pixel-Reasoner~\cite{wang2025pixelreasonerincentivizingpixelspace}
& Qwen2.5-VL-7B
& 25.70 & 28.34 & 30.56 & 28.73
& 53.23 & 23.00 & 62.93 & 46.58
& 75.86 & 45.17 & 50.41 & 55.26
& 30.69 & \zjRankThree{65.30} & 34.75
& 40.58 & 41.18 \\

\midrule

GeoChat~\cite{kuckreja2024geochat}
& Vicuna-7B
& 20.64 & 22.60 & 20.13 & 20.65
& 22.87 & 16.88 & 26.25 & 22.04
& 66.19 & 34.58 & 41.25 & 45.64
& 27.82 & 24.79 & 28.60
& 22.63 & 27.91 \\

VHM~\cite{pang2024vhmversatilehonestvision}
& Vicuna-7B
& 13.86 & 17.04 & 13.88 & 14.33
& 23.11 & 16.97 & 29.44 & 23.22
& 60.52 & 27.61 & 35.87 & 39.82
& 30.77 & 35.23 & 31.72
& 23.41 & 26.50 \\

RSUniVLM~\cite{liu2024rsunivlmunifiedvisionlanguage}
& Qwen2-0.5B
& 23.94 & 28.63 & 29.13 & 27.45
& 24.38 & 23.90 & 28.16 & 25.49
& 63.57 & 34.41 & 38.20 & 43.37
& 24.87 & 49.96 & 24.87
& 26.79 & 29.60 \\

SkySenseGPT~\cite{luo2024skysensegpt}
& Vicuna-7B
& 23.28 & 22.03 & 24.76 & 23.91
& 22.87 & 16.88 & 26.25 & 22.04
& 66.49 & 38.24 & 39.71 & 45.77
& 29.04 & 28.03 & 29.72
& 22.99 & 28.89 \\

TEOChat~\cite{irvin2024teochat}
& Vicuna-7B
& 18.40 & 28.34 & 22.83 & 22.26
& 22.95 & 16.88 & 26.41 & 22.12
& 64.13 & 32.71 & 42.84 & 45.54
& 25.23 & 33.26 & 26.60
& 23.90 & 28.08 \\

EarthDial~\cite{soni2024earthdial}
& Phi-3-mini
& 15.32 & 25.80 & 19.60 & 19.17
& 26.53 & 25.86 & 42.56 & 31.70
& 63.83 & 29.62 & 38.90 & 42.68
& 30.41 & 43.35 & 32.32
& 34.29 & 32.03 \\

GeoLLaVA-8K~\cite{wang2025geollava}
& Qwen2-7B
& 3.39 & 3.11 & 3.33 & 3.31
& 32.27 & 22.27 & 28.08 & 27.58
& 10.70 & 4.83 & 3.17 & 5.36
& 29.66 & 54.05 & 31.63
& 43.47 & 22.27 \\

Coarse-to-Fine~\cite{Luo_2025_ICCV}
& Qwen2-7B
& 22.62 & \zjRankTwo{35.40} & 26.56 & 26.62
& 41.75 & 33.61 & 50.28 & 41.95
& 71.35 & 35.32 & 36.52 & 44.57
& 32.90 & 49.73 & 33.28
& 33.90 & 36.06 \\

\midrule

\textbf{EarthJev}
& Qwen3.5-9B
& \zjRankOne{30.63} & \zjRankOne{36.53} & \zjRankOne{36.68} & \zjRankOne{34.79}
& \zjRankOne{72.03} & \zjRankOne{39.89} & \zjRankOne{77.01} & \zjRankOne{63.16}
& 77.10 & \zjRankTwo{52.74} & 54.07 & \zjRankFour{59.27}
& \zjRankOne{53.50} & 60.53 & \zjRankOne{52.25}
& \zjRankOne{55.81} & \zjRankOne{53.06} \\

\bottomrule
\end{tabular}%
}

\end{table*}

%% file: table/performance_XLRS-Bench-lite.tex
\begin{table*}[!t]
\centering
\caption{Performance comparison of Earth with existing methods on XLRS-Bench-lite, covering 13 fine-grained Level-3 (L3) task categories. Training settings are indicated by \ding{51} (trained), \hmark (answering model frozen, with only lightweight decision modules trained), and \ding{55} (training-free).}
\label{tab:main2}
\fontsize{7.5}{9}\selectfont
\setlength{\tabcolsep}{2pt}
\renewcommand{\arraystretch}{1.0}
\resizebox{\textwidth}{!}{%
\begin{tabular}{llc*{14}{c}}
\toprule
Method & Backbone & Training & AD & CCR & ECR & OC & OCC & OCL & OLUC & OMS & OSR & RC & RCCD & RLUC & RP & Avg. \\
\midrule
ZoomSearch & Qwen3.5-9B & \ding{55} & \zjRankOne{77.00} & 49.00 & 79.00 & \zjRankTwo{33.33} & 39.50 & 41.12 & 26.00 & 46.67 & \zjRankFour{35.00} & \zjRankOne{58.00} & 30.00 & \zjRankFour{84.50} & 26.00 & 44.48 \\
GazeEarth & Qwen3-VL-8B & \ding{55} & \zjRankTwo{76.00} & \zjRankOne{63.00} & \zjRankOne{86.00} & \zjRankTwo{33.33} & \zjRankThree{52.62} & \zjRankThree{50.12} & 31.00 & \zjRankTwo{71.67} & \zjRankOne{37.80} & 51.00 & \zjRankOne{53.33} & \zjRankFour{84.50} & \zjRankThree{49.00} & \zjRankThree{52.95} \\
ViCrop & Qwen3-VL-8B & \ding{55} & 70.00 & \zjRankFour{52.00} & 79.00 & 16.67 & 43.25 & 40.00 & 19.00 & 65.00 & 28.40 & 44.00 & \zjRankFour{45.00} & 77.50 & 37.00 & 43.51 \\
AttWarp & Qwen3-VL-8B & \ding{55} & \zjRankThree{74.00} & 46.00 & 79.00 & 16.67 & 46.25 & \zjRankFour{42.00} & 23.00 & 66.67 & 31.00 & 44.00 & \zjRankThree{46.67} & 76.50 & 39.00 & 45.36 \\
WeaveEarth & Qwen3-VL-8B & \ding{55} & \zjRankThree{74.00} & \zjRankTwo{56.00} & 78.00 & 21.67 & 36.00 & 37.62 & 15.00 & 66.67 & 28.00 & 51.00 & \zjRankFour{45.00} & 83.00 & 25.00 & 41.36 \\
Zoom-Refine & Qwen3.5-9B & \ding{55} & 68.00 & 49.00 & \zjRankFour{81.00} & \zjRankOne{35.00} & 41.88 & 39.25 & 32.00 & \zjRankTwo{71.67} & 32.80 & \zjRankThree{53.00} & 33.33 & \zjRankTwo{85.50} & 41.00 & 45.19 \\
\midrule
RS-OPSD & Qwen3-VL-8B & \ding{51} & \zjRankTwo{76.00} & 48.00 & \zjRankTwo{83.00} & \zjRankFour{28.33} & 46.75 & 40.88 & \zjRankThree{36.00} & \zjRankThree{70.00} & 28.00 & \zjRankFour{52.00} & \zjRankThree{46.67} & \zjRankThree{85.00} & \zjRankThree{49.00} & \zjRankFour{46.82} \\
GeoLens & Qwen2.5-VL-7B & \ding{51} & 72.00 & \zjRankThree{53.00} & \zjRankThree{82.00} & \zjRankThree{30.00} & \zjRankFour{48.88} & 32.75 & 1.00 & 66.67 & 30.60 & 39.00 & \zjRankTwo{50.00} & 81.00 & \zjRankFour{48.00} & 43.86 \\
GeoEyes & Qwen2.5-VL-7B & \ding{51} & \zjRankThree{74.00} & 40.00 & \zjRankFour{81.00} & 26.67 & \zjRankTwo{59.00} & \zjRankOne{68.00} & 19.00 & \zjRankThree{70.00} & 32.20 & 38.00 & 33.33 & 76.00 & \zjRankTwo{56.00} & \zjRankTwo{55.68} \\
ZoomEarth & Qwen2.5-VL-3B & \ding{51} & 61.00 & 1.00 & 69.00 & 10.00 & 24.38 & 15.62 & 3.00 & 31.67 & 12.60 & 2.00 & 10.00 & 53.00 & 44.00 & 22.73 \\
\midrule
DeepEyes & Qwen2.5-VL-7B & \ding{51} & 70.00 & 40.00 & 76.00 & 16.67 & 41.88 & 35.75 & \zjRankTwo{43.00} & \zjRankFour{68.33} & 30.60 & 39.00 & \zjRankThree{46.67} & 77.00 & 40.00 & 42.69 \\
Mini-o3 & Qwen2.5-VL-7B & \ding{51} & 71.00 & 17.00 & 72.00 & 10.00 & 22.00 & 30.00 & \zjRankFour{33.00} & 26.67 & 24.40 & 41.00 & 1.67 & 80.00 & 16.00 & 31.53 \\
Pixel-Reasoner & Qwen2.5-VL-7B & \ding{51} & 65.00 & 48.00 & 76.00 & 18.33 & 36.50 & 34.75 & 16.00 & 63.33 & 34.40 & 41.00 & 43.33 & 75.50 & 36.00 & 40.58 \\
\midrule
GeoChat & Vicuna-7B & \ding{51} & 36.00 & 21.00 & 48.00 & 16.67 & 21.12 & 16.75 & 1.00 & 35.00 & 23.80 & 29.00 & 28.33 & 23.00 & 46.00 & 22.63 \\
VHM & Vicuna-7B & \ding{51} & 34.00 & 21.00 & 48.00 & 16.67 & 20.50 & 16.75 & 9.00 & 35.00 & 28.20 & 30.00 & 28.33 & 23.00 & 46.00 & 23.41 \\
RSUniVLM & Qwen2-0.5B & \ding{51} & 27.00 & 32.00 & 21.00 & 20.00 & 30.88 & 24.62 & 2.00 & 35.00 & 27.80 & 23.00 & 18.33 & 36.50 & 20.00 & 26.79 \\
SkySenseGPT & Vicuna-7B & \ding{51} & 37.00 & 21.00 & 48.00 & 16.67 & 22.12 & 16.75 & 1.00 & 35.00 & 23.80 & 29.00 & 28.33 & 24.00 & 46.00 & 22.99 \\
TEOChat & Vicuna-7B & \ding{51} & 37.00 & 21.00 & 49.00 & 16.67 & 23.38 & 16.75 & 2.00 & 40.00 & 25.00 & 29.00 & 28.33 & 27.50 & 46.00 & 23.90 \\
EarthDial & Phi-3-mini & \ding{51} & 64.00 & 44.00 & 69.00 & 21.67 & 33.75 & 31.13 & 2.00 & 55.00 & 25.40 & 43.00 & \zjRankFour{45.00} & 41.50 & 32.00 & 34.29 \\
GeoLLaVA-8K & Qwen2-7B & \ding{51} & 68.00 & 50.00 & 77.00 & 25.00 & 42.00 & 31.75 & \zjRankOne{51.00} & 65.00 & \zjRankThree{35.20} & 41.00 & \zjRankTwo{50.00} & 68.00 & \zjRankOne{66.00} & 43.47 \\
Coarse-to-Fine & Qwen2-7B & \ding{51} & \zjRankFour{73.00} & 45.00 & 75.00 & 23.33 & 29.38 & 32.50 & 6.00 & 26.67 & 25.20 & 43.00 & \zjRankTwo{50.00} & 47.50 & 26.00 & 33.90 \\
\midrule
\textbf{EarthJev} & Qwen3.5-9B & \hmark & \zjRankTwo{76.00} & 51.00 & 77.00 & 26.67 & \zjRankOne{61.88} & \zjRankTwo{59.13} & 11.00 & \zjRankOne{80.00} & \zjRankTwo{37.60} & \zjRankTwo{54.00} & 41.67 & \zjRankOne{88.00} & 29.00 & \zjRankOne{55.81} \\
\bottomrule
\end{tabular}
}
\par\vspace{4pt}
\end{table*}

%% file: table/ablation_study.tex
\begin{table*}[!t]
\centering
\caption{Component ablation of EarthJev with Qwen3.5-9B.
Results are reported as accuracy (\%).
The full configuration is shaded in light coral;
the best result in each column is boldfaced.}
\label{tab:component_ablation}

\begingroup
\fontsize{9}{11}\selectfont
\setlength{\tabcolsep}{5pt}
\renewcommand{\arraystretch}{1.22}

\begin{tabular}{@{}cccccccccc@{}}
\toprule
\rowcolor{zjHeader}
\textcolor{zjInk}{\textbf{Group}} &
\textcolor{zjInk}{\textbf{EDM}} &
\textcolor{zjInk}{\textbf{SPV}} &
\textcolor{zjInk}{\textbf{CAS}} &
\textcolor{zjInk}{\textbf{MME-RS}} &
\textcolor{zjInk}{\textbf{LRS-VQA}} &
\textcolor{zjInk}{\textbf{XLRS-lite}} &
\textcolor{zjInk}{\textbf{LRS-GRO}} &
\textcolor{zjInk}{\textbf{RSHR-Bench}} &
\textcolor{zjInk}{\textbf{Avg.}} \\
\midrule

1 & \ding{55} & \ding{55} & \ding{55}
& 60.62 & 32.46 & 44.48
& 53.91 & 37.52 & 45.80 \\

2 & \ding{51} & \ding{55} & \ding{55}
& 62.97 & 34.52 & 44.29
& 55.42 & 38.99 & 47.24 \\

3 & \ding{51} & \ding{51} & \ding{55}
& 62.81 & \textbf{35.20} & 55.00
& 56.04 & \textbf{54.16} & 52.64 \\

\rowcolor{zjFull}
4 & \ding{51} & \ding{51} & \ding{51}
& \textbf{63.16} & 34.79 & \textbf{55.81}
& \textbf{59.27} & 52.25 & \textbf{53.06} \\

\bottomrule
\end{tabular}
\endgroup
\end{table*}

\begin{figure}[htbp]
    \centering
    \includegraphics[width=0.99\columnwidth]{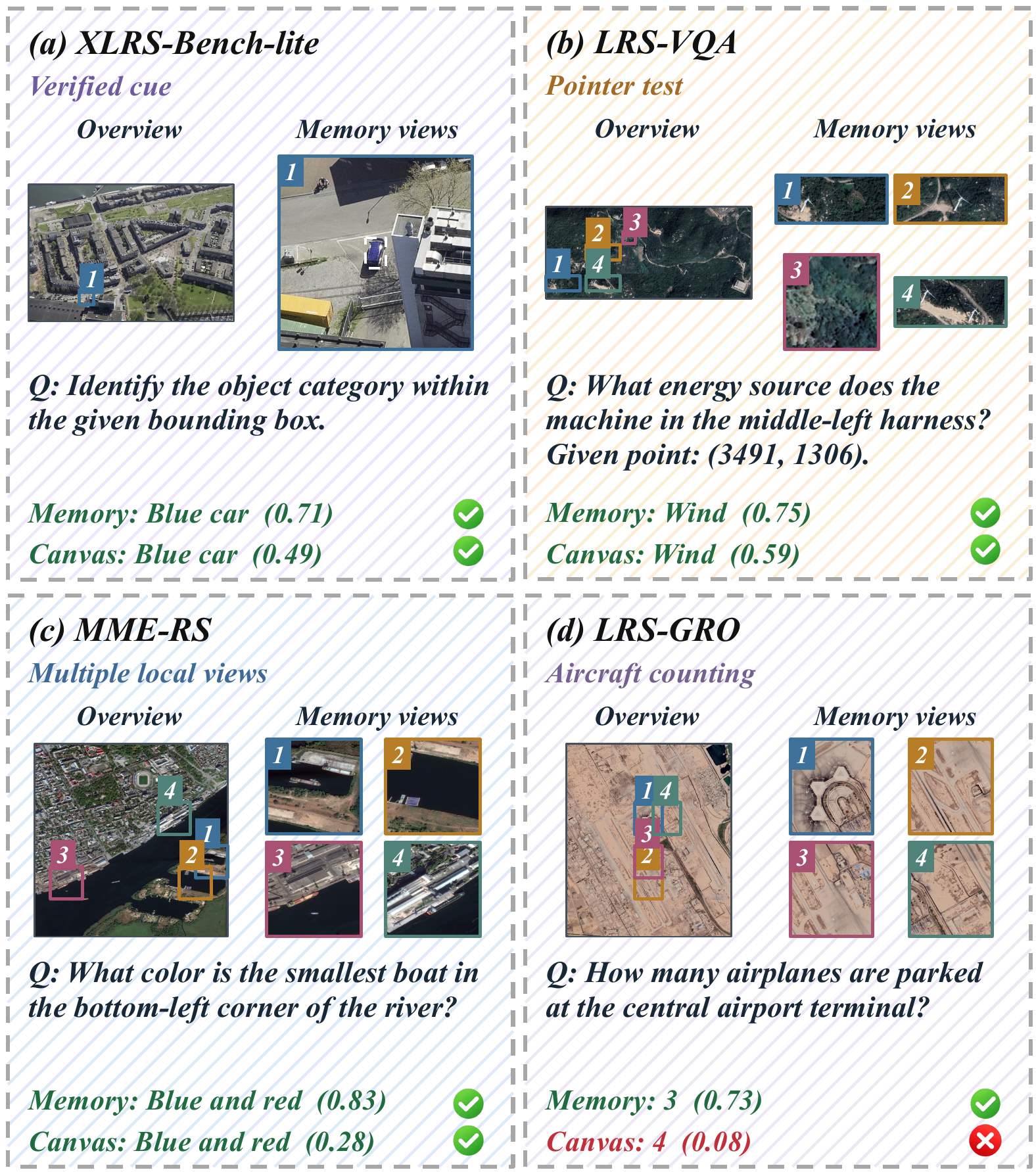}
    \caption{Visual memory examples combining a global overview
    with local views indexed by their source locations.
    Memory and canvas answers are shown with calibrated confidence
    in parentheses.}
    \label{fig:visual_memory}
\end{figure}

%% file: table/cross_model_generalization.tex
\begin{table}[!t]
\centering
\caption{Performance of EarthJev across answering models (accuracy, \%). Gray and coral cells show the base and EarthJev-augmented results, respectively; blue cells report gains in percentage points.}
\label{tab:answering_models}

\begingroup
\fontsize{7.5}{9}\selectfont
\setlength{\tabcolsep}{2pt}
\renewcommand{\arraystretch}{1.2}

\resizebox{\columnwidth}{!}{%
\begin{tabular}{llcccccc}
\toprule
\rowcolor{modelHeader}
\textbf{Answerer} & \textbf{Setting}
& \textbf{MME-RS} & \textbf{LRS-VQA} & \textbf{XLRS-lite}
& \textbf{LRS-GRO} & \textbf{RSHR-Bench} & \textbf{Avg.} \\
\midrule

\multirow{3}{*}{Qwen3.5-4B}
& \modelbasecell{Base}
& \modelbasecell{51.95} & \modelbasecell{29.74}
& \modelbasecell{41.95} & \modelbasecell{58.11}
& \modelbasecell{37.52} & \modelbasecell{43.85} \\

& \modelourscell{+\,EarthJev}
& \modelourscell{57.76} & \modelourscell{30.32}
& \modelourscell{53.96} & \modelourscell{59.01}
& \modelourscell{50.87} & \modelourscell{\textbf{50.38}} \\

& \gaincell{Gain (pp)}
& \gaincell{+5.81} & \gaincell{+0.57}
& \gaincell{+12.01} & \gaincell{+0.90}
& \gaincell{+13.34} & \gaincell{\textbf{+6.53}} \\
\midrule

\multirow{3}{*}{Qwen3-VL-8B}
& \modelbasecell{Base}
& \modelbasecell{55.97} & \modelbasecell{32.27}
& \modelbasecell{46.10} & \modelbasecell{59.27}
& \modelbasecell{36.05} & \modelbasecell{45.93} \\

& \modelourscell{+\,EarthJev}
& \modelourscell{56.39} & \modelourscell{34.97}
& \modelourscell{54.38} & \modelourscell{59.40}
& \modelourscell{49.48} & \modelourscell{\textbf{50.92}} \\

& \gaincell{Gain (pp)}
& \gaincell{+0.43} & \gaincell{+2.70}
& \gaincell{+8.28} & \gaincell{+0.13}
& \gaincell{+13.43} & \gaincell{\textbf{+4.99}} \\
\midrule

\multirow{3}{*}{Intern-S1-mini}
& \modelbasecell{Base}
& \modelbasecell{43.37} & \modelbasecell{28.95}
& \modelbasecell{38.70} & \modelbasecell{53.21}
& \modelbasecell{36.48} & \modelbasecell{40.14} \\

& \modelourscell{+\,EarthJev}
& \modelourscell{47.62} & \modelourscell{30.89}
& \modelourscell{44.87} & \modelourscell{54.10}
& \modelourscell{40.29} & \modelourscell{\textbf{43.55}} \\

& \gaincell{Gain (pp)}
& \gaincell{+4.25} & \gaincell{+1.94}
& \gaincell{+6.17} & \gaincell{+0.89}
& \gaincell{+3.81} & \gaincell{\textbf{+3.41}} \\
\midrule

\multirow{3}{*}{\shortstack[l]{LLaVA-\\OneVision-2-8B}}
& \modelbasecell{Base}
& \modelbasecell{57.97} & \modelbasecell{19.69}
& \modelbasecell{39.84} & \modelbasecell{39.74}
& \modelbasecell{35.27} & \modelbasecell{38.50} \\

& \modelourscell{+\,EarthJev}
& \modelourscell{59.02} & \modelourscell{28.98}
& \modelourscell{42.21} & \modelourscell{58.33}
& \modelourscell{36.57} & \modelourscell{\textbf{45.02}} \\

& \gaincell{Gain (pp)}
& \gaincell{+1.04} & \gaincell{+9.29}
& \gaincell{+2.37} & \gaincell{+18.59}
& \gaincell{+1.30} & \gaincell{\textbf{+6.52}} \\

\bottomrule
\end{tabular}%
}
\endgroup
\end{table}

%% file: table/efficiency.tex
\begin{table}[!t]
\centering
\caption{Inference cost under the same answering model. s/q: seconds per question on one NVIDIA H100 GPU}
\label{tab:efficiency}
\begingroup
\fontsize{7.5}{9}\selectfont
\setlength{\tabcolsep}{10pt}
\renewcommand{\arraystretch}{1.15}
\begin{tabular}{lccc}
\toprule
\textbf{Method} & \textbf{s/q} & \textbf{Rel. time} & \textbf{Avg.} \\
\midrule
\multicolumn{4}{l}{\textit{Qwen3.5-9B}} \\
\quad Direct & 3.30 & 1.0$\times$ & 46.93 \\
\quad ZoomSearch & 11.30 & 3.4$\times$ & 45.80 \\
\quad Zoom-Refine & 11.34 & 3.4$\times$ & 46.94 \\
\rowcolor{zjFull}
\quad EarthJev & 1.67 & 0.5$\times$ & 53.06 \\
\midrule
\multicolumn{4}{l}{\textit{Qwen3-VL-8B}} \\
\quad Direct & 3.41 & 1.0$\times$ & 47.31 \\
\quad ZoomSearch & 9.86 & 2.9$\times$ & 41.86 \\
\quad ViCrop & 2.01 & 0.6$\times$ & 40.93 \\
\quad AttWarp & 1.98 & 0.6$\times$ & 42.99 \\
\quad WeaveEarth & 1.92 & 0.6$\times$ & 43.10 \\
\quad GazeEarth & 3.88 & 1.1$\times$ & 48.36 \\
\quad RS-OPSD & 3.48 & 1.0$\times$ & 48.07 \\
\rowcolor{zjFull}
\quad EarthJev & 1.68 & 0.5$\times$ & 50.92 \\
\bottomrule
\end{tabular}

\endgroup
\end{table}